# PhaseSync-Exo: Human Clock Anchored Reference Adaptation for Dynamic Gait Tracking

Kaijie Qi[1], Yuehan Wang[1], Kaiming Xu[1], Chong Li[1], Jiakuo Yu[1]

[1]Tsinghua University

**Abstract—Human-aware exoskeleton walking requires reconstructing gait, tracking diverse motions under dynamic constraints, and preserving human timing. We present PhaseSync-Exo, which combines two-IMU CNN–Transformer reconstruction, factorized amplitude–cadence retargeting with curriculum-trained recurrent control, and a human-clock-anchored adapter (HCA). HCA combines human-clock attraction with robot-relative feedback to adjust reference rate while preserving forward progression and continuity. The reconstruction module achieves 3.24° MAE on a held-out recording, and the frozen tracking policy completes 356/357 amplitude–frequency trials. Two complementary dynamic comparisons isolate HCA's timing benefit without retraining. Against robot-relative correction, HCA reduces human-clock phase MAE from 95.55° to 10.99°, limiting reference drift. When human and delivered phases initially differ, it reduces phase MAE from 72.02° to 13.90° versus fixed-clock continuation. Compared with immediate phase reset, HCA reduces transient reference-tracking hip RMSE by 22.7% without reference jumps. All 420 timing rollouts complete without falls. These simulation results support continuous phase acquisition and sustained alignment to an independent human clock.**

## I. Introduction

Powered lower-limb exoskeletons must satisfy two coupled requirements: dynamically stable locomotion and motion coordination with a periodically moving human. A controller that merely keeps the mechanism upright may oppose the wearer when assistance is delivered at the wrong phase. Conversely, direct playback of a human joint trajectory may violate contact, actuator, and balance constraints. The central problem is therefore to preserve a human gait's timing and dominant hip–knee pattern while allowing the robot to modify dynamically critical degrees of freedom.

Physics-based imitation methods such as DeepMimic [1], adversarial motion priors [2], and BeyondMimic-style tracking [3] convert kinematic demonstrations into dynamically feasible policies; retargeting quality also affects tracking [4]. Temporal adaptation poses a separate problem. Fixed-clock (FC) continuation preserves cadence but leaves an initial phase offset unchanged. Robot-relative correction (RRC) accommodates robot timing, yet can accumulate drift without attraction to the independent human clock. Immediate phase reset removes clock mismatch by introducing a reference discontinuity. The question is how to recover human timing continuously while keeping the reference dynamically trackable.

PhaseSync-Exo separates motion diversity, temporal alignment, and dynamic control. Exact-forward-kinematics retargeting builds a parameterized robot-motion family. We distinguish an amplitude–cadence curriculum (ACC), which varies gait shape and cadence jointly, from a wide-frequency curriculum (WFC), which expands cadence coverage at nominal amplitude. These descriptive names identify the training configurations throughout this paper. A human-clock-anchored circular phase loop then adjusts reference-query rate, while the recurrent policy realizes the selected motion under physical constraints.

The hierarchy has three explicit interfaces. The motion descriptor selects the desired amplitude–frequency condition, the phase synchronizer determines the current and preview frames, and the GRU–PPO controller realizes those frames through constrained dynamics. This modularity makes each failure attributable to reference coverage, temporal alignment, or physical tracking. It also keeps the actor deployable because simulator-only state is confined to the asymmetric critic during training.

The contributions of this manuscript are threefold:

1) A causal two-IMU CNN–Transformer for reconstructing continuous hip, knee, and ankle kinematics as human motion input. Evaluation on a held-out recording separates reconstruction performance from the downstream controller's tracking accuracy.

2) A factorized multi-trajectory training framework that separates joint excursion from cadence and couples exact-FK retargeting with recurrent asymmetric actor–critic control. Complementary curricula support joint amplitude–cadence training and wide-frequency training; the latter is evaluated for transfer to unseen amplitudes without policy changes.

3) A human-clock-anchored adapter that combines circular human and robot phase errors within a positive bounded reference-rate update. Frozen-policy dynamic comparisons establish drift suppression relative to RRC and continuous phase acquisition relative to FC continuation, with lower transient tracking error than immediate phase reset.

## II. Related Work

### A. Physics-Based Motion Imitation

DeepMimic combines task and imitation objectives for physics-based skill learning [1]. AMP learns a motion prior from demonstrations [2], and BeyondMimic develops high-fidelity tracking and policy composition [3], [4]. These approaches establish how motion data can supervise dynamics. PhaseSync-Exo addresses a different gap: it parameterizes a retargeted motion family and closes the timing loop around that family, so reference diversity and synchronization remain explicit rather than implicit in a trajectory-specific policy.

### B. Reinforcement Learning for Legged and Exoskeleton Control

PPO provides a clipped surrogate objective for stable policy updates [5], and generalized advantage estimation controls the bias–variance trade-off of the policy-gradient target [6]. Parallel simulation and dynamics randomization support transfer [7], [8]. In PhaseSync-Exo, reinforcement learning is responsible for dynamics realization, not for defining the desired human frame. The synchronized reference enters the actor as a command, while a privileged critic supplies clean simulator state only during training. This separation preserves an interpretable timing mechanism outside the learned controller.

### C. Gait Phase and Human–Robot Coordination

Gait phase is a circular progress variable rather than an absolute timestamp. Adaptive oscillators and phase-variable controllers synchronize assistance with periodic motion [9], [10], [11], [12], while interaction-torque adaptation addresses human–robot disagreement [13]. A scalar phase estimate alone does not specify which member of a motion family should be tracked, and direct nearest-frame correction can violate temporal continuity. PhaseSync-Exo couples circular phase comparison to a parameterized retargeted motion bank and constrains correction at the cursor-rate level.

## III. Design of the Proposed Exoskeleton

The exoskeleton has six actuated sagittal joints ordered as left hip, right hip, left knee, right knee, left ankle, and right ankle, which is shown in Figure 1. A retargeted library $M$ contains joint position $q^*$, joint velocity $\dot{q}^*$, torso pose $R^*$, and maximum-coordinate states of seven tracked bodies at 60 Hz. The descriptor $\theta = [s_\omega, s_h, s_k, s_a]$ specifies cadence and hip, knee, and ankle amplitude scales. The policy generates bounded actions $a \in [-1,1]^6$ subject to contact, actuator, and balance constraints.

## IV. Human Gait Kinematics Reconstruction

Two synchronized IMUs supply 18 channels: three-axis acceleration, angular velocity, and orientation from each sensor. Two IMUs are respectively installed on the thigh and shank of the exoskeleton. A CNN–Transformer processes a trailing 32-sample window (approximately 1.62 s) to estimate sagittal hip, knee, and ankle angles. The module reconstructs joint kinematics for downstream reference generation using recordings without activity or phase labels. The workflow is shown in Figure 2.

## V. Factorized Multi Trajectory Tracking

Joint angles are scaled about the upright neutral pose in three bilateral groups at 60 Hz: hip, knee, and ankle. For each amplitude variant, root forward displacement is scaled by 0.6 times the hip scale plus 0.4 times the knee scale. Exact robot forward kinematics recomputes the seven tracked body poses; a smoothed, bounded root-height correction preserves the source foot-height profile. Velocities are then recomputed. Exact FK provides geometric consistency between joint and body references; the recurrent policy realizes these references under simulated dynamics.

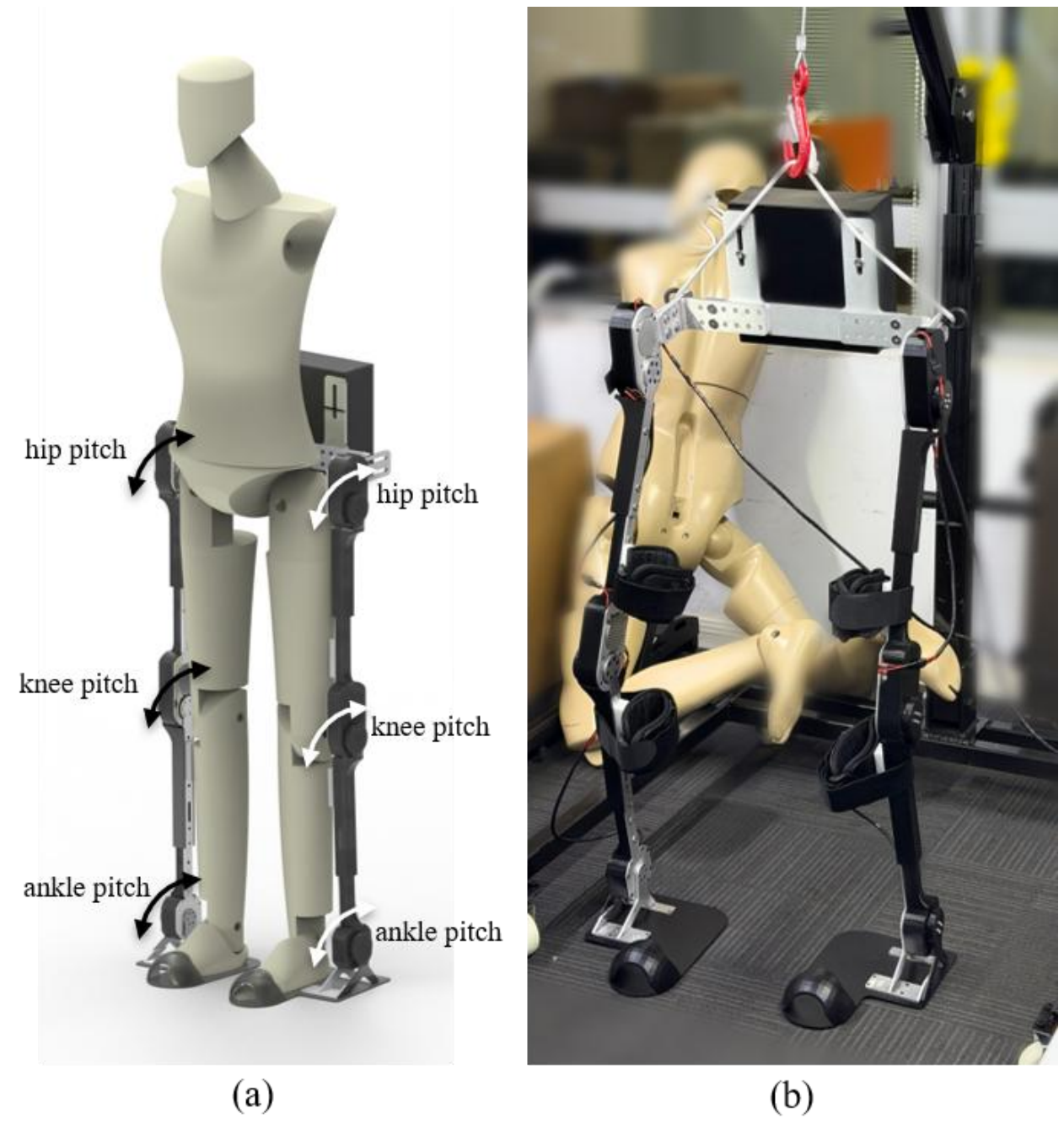

Figure 1. (a) Exoskeleton model and (b) prototype.

In order to enable the exoskeleton to effectively track different amplitudes of gait, we established multiple sets of trajectories. The group contains 17 amplitude variants: IDs 0–12 are ACC training references and IDs 13–16 are held out from ACC. Hip, knee, and ankle scale ranges are 0.85–1.15, 0.80–1.20, and 0.90–1.10. ACC expands variation from 15% to full range over 5,000 PPO iterations, with continuously sampled cadence scales of 0.75–1.25. WFC instead fixes amplitude to nominal variant 0 and expands cadence from 10% to full range over 10,000 iterations, reaching 0.50–1.60.

Amplitude scales joint excursion about the neutral pose; cadence scales traversal speed without changing that pose path. Current and two preview joint states condition one recurrent actor (shown in Figure 3. ). Interpolated positions and reference velocities share the corrected cursor rate. This factorized representation supports amplitude–frequency combinations without storing a separate clip for every time scale.

Figure 3. illustrates how the factorized references are realized by a dynamics-aware tracking policy. At each 60-Hz control step, the retargeted motion bank supplies current joint position and velocity targets, two future reference previews, and a gait descriptor containing phase, cadence, amplitude scales, and desired speed. These features are combined with torso-IMU signals, six joint encoder readings, and the previous action to form a 70-dimensional actor observation. A 256-unit GRU followed by a 512–256–128 MLP generates six bounded residual joint-position commands, which are executed through PD control under motor torque–speed limits. A separate 139-dimensional critic uses privileged root and body states available

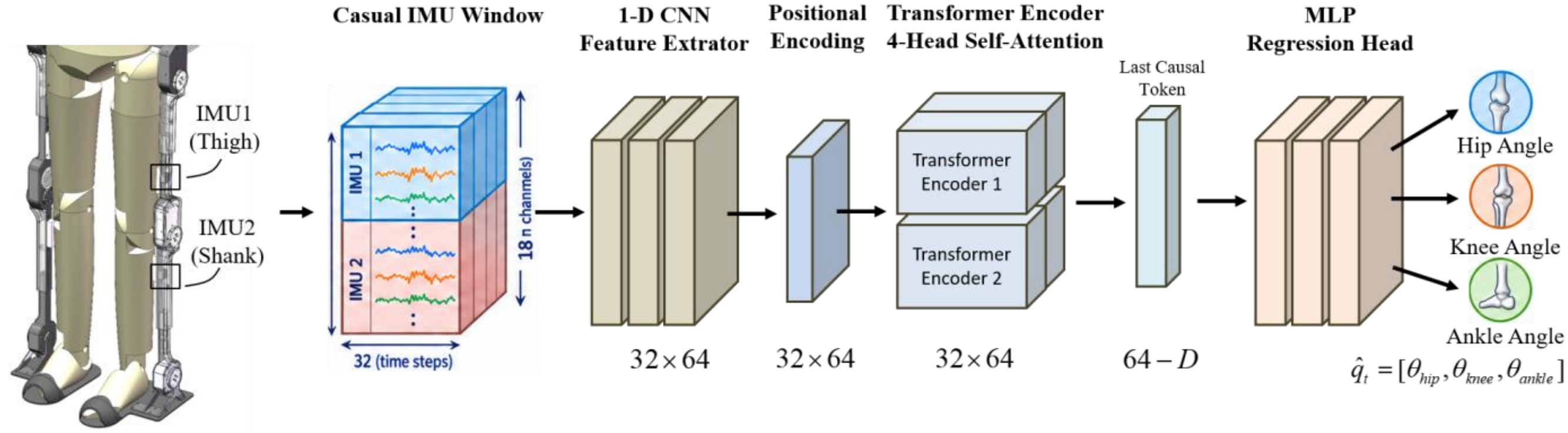

Figure 2. Workflow of the proposed gait kinematics reconstruction.

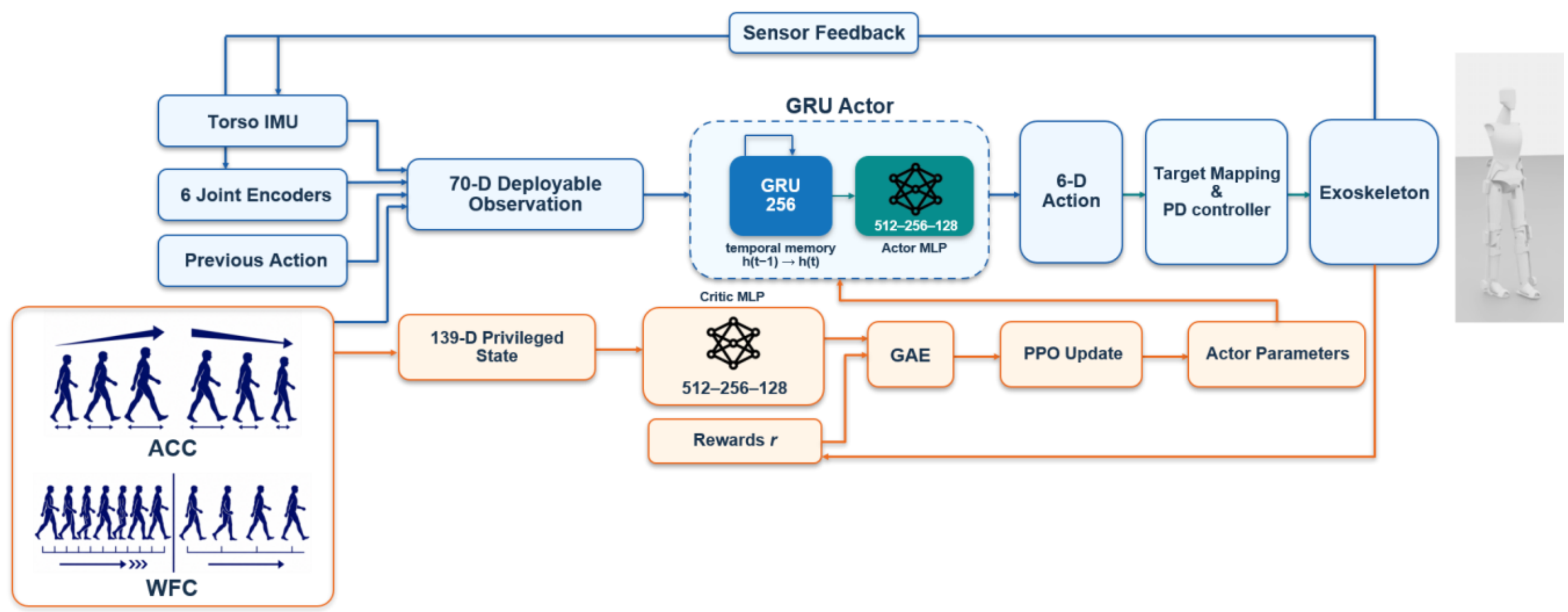

Figure 3. Workflow of multi trajectory tracking training.

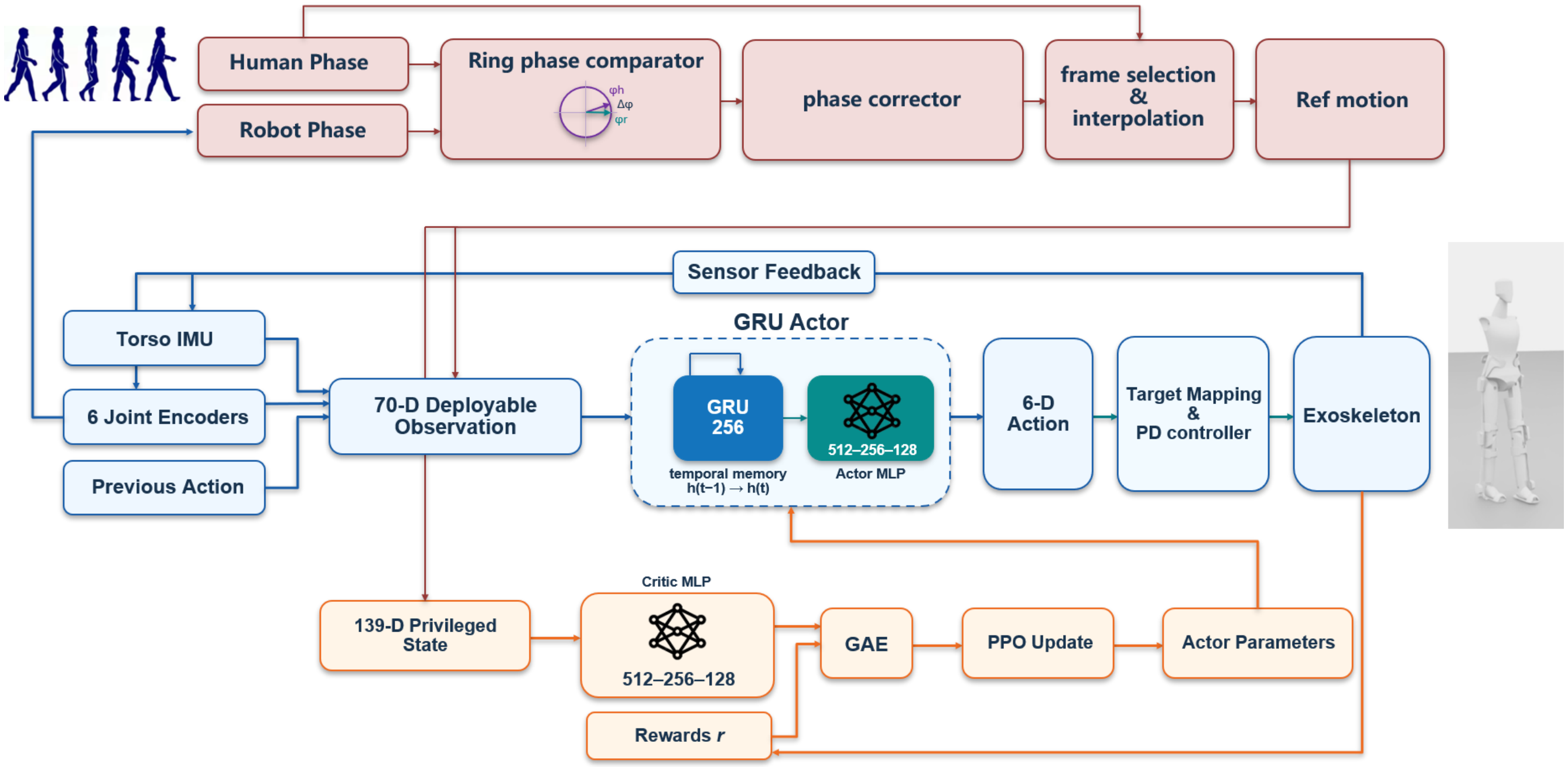

Figure 4. Workflow of human-clock-anchored reference adaptation.

in simulation. PPO with generalized advantage estimation updates both networks using motion-tracking, stability, and actuator-related rewards. The amplitude–cadence curriculum varies both gait factors during training; the wide-frequency curriculum trains at nominal amplitude and evaluates amplitude transfer across the 17-trajectory bank. During deployment, only the actor and joint controller are required

## VI. PhaseSync Exo Method

### A. Reference Phase and Synchronization Interface

We express phase in cycles and distinguish the human phase $\phi_h$, the delivered reference phase $\phi_d$, and the estimated robot phase $\hat{\phi}_r$. The human phase $\phi_h$ is estimated online from wearable sensing and serves as an independent timing signal for reference synchronization. The delivered reference cursor is updated by the bounded phase adapter, while the robot phase is estimated from proprioceptive joint measurements. The motion bank contains 62 frames per cycle.

### B. Proprioceptive Robot Phase Estimate

The encoder estimator matches joint positions and velocities over 29 candidates within ±0.22 cycle of the delivered cursor. Hip/knee/ankle weights are 1.0/1.25/0.5, with 0.30 rad/2.50 rad/s normalization. The winning offset, normalized by cycle length, is filtered with EMA coefficient 0.2. Local unwrapped queries handle cycle boundaries within this search neighborhood.

### C. Human Clock Anchored Bounded Phase Correction

Let $\bar{e}$ denote the filtered robot-minus-delivered phase error in cycles, $s_\omega$ the supplied cadence multiplier, $f_s$ the frame rate, and $u$ the unwrapped cursor. The original RRC uses gain 1.5 and the common cursor update:

$$\delta_t = clip(1.5\overline{e}_t, -b, b) \tag{1}$$

$$u_{t+1} = u_t + f_s \Delta t s_\omega (1+\delta_t) \tag{2}$$

Here $b$=0.25 in ACC and the kinematic test, and 0.10 in WFC dynamics. RRC integrates robot-relative timing corrections without an attraction to human timing. A persistent nonzero correction can therefore accumulate human-clock drift despite strictly positive reference progression.

Anchored PhaseSync adds a separate circular human-minus-delivered error. With wrap returning [−0.5, 0.5) cycle, the two errors and correction are shown below:

$$e_h = wrap(\varphi_h - \varphi_d) \tag{3}$$

$$e_r = wrap(\hat{\varphi}_r - \varphi_d) \tag{4}$$

$$\delta = clip_{[-b,.b]}(k_h e_h + k_r \overline{e}_r) \tag{5}$$

Human/robot gains are fixed at 6/1.5 with $b$ = 0.10. Delivered frequency equals human frequency times (1+$\delta$); continuous-time gains equal these dimensionless gains times human frequency. Robot lag slows the reference, while human-clock attraction pulls it back. The policy and robot estimator are unchanged.

Positive human cadence and $b$<1 preserve forward progression. Locally, away from wrap and saturation, human-reference error has a restoring term and filtered robot error acts as a disturbance. Constant bias leaves a finite offset.

We further evaluate phase-based reference adaptation under temporary changes in actuator capability. The experiments measure tracking, human-clock error, and recovery with the same policy and adapter gains.

### D. Synchronized Frame Query

The synchronized cursor selects adjacent frames of the chosen motion variant. Joint positions are linearly interpolated, while body quaternions use shortest-path normalized interpolation. Reference velocities are multiplied by the effective playback scale, consistently with cursor advancement:

$$q_{ref} = s_\omega (1+\delta_t)\dot{q}_{lib}(u_t) \tag{6}$$

Preview frames at +2 and +5 policy steps use the same corrected rate. Thus the actor receives temporally consistent current and future commands instead of poses written directly into the simulator. The recurrent policy must still realize these commands under contact, balance, and motor constraints.

### E. Multi Reference Recurrent Asymmetric Actor Critic

Both configurations use a 70-D actor observation. The 39-D base interface contains reference joint positions/velocities (12-D), reference-to-torso orientation (6-D), torso angular velocity (3-D), encoder positions/velocities (12-D), and previous action (6-D). Two preview position/velocity pairs add 24 dimensions; phase sine/cosine, cadence, three amplitude scales, and desired speed add seven. A single-layer, 256-unit GRU feeds ELU layers of width 512–256–128 and a Gaussian six-action head. The 139-D privileged critic additionally uses clean simulator state and is not deployed.

### F. Action Mapping Motor Limits and Stability Objectives

Actor outputs are clipped and mapped to residual position targets around the default pose:

$$q_{des} = clip(q_0 + 0.5a, q_{\min}, q_{\max}) \tag{7}$$

Targets update at 60 Hz and are tracked by joint PD control at 240 Hz. Hip and knee modules use 10 N·m rated and 36 N·m stall torque, with 50/60 r/min rated/no-load speed; ankle modules use 3.5/12 N·m and 250/270 r/min. The torque–speed envelope is enforced during training. The reward combines reference body position, orientation, linear velocity, and angular velocity with direct joint tracking, action-rate, saturation, joint-limit, contact, torque, and power terms. Hip and knee receive greater weight in phase matching, but all six joints contribute to balance.

## VII. Training and Evaluation Design

### A. Simulation and Curriculum Training

Isaac Sim/Isaac Lab training uses 1024 environments, 240 Hz physics, 60 Hz control, and 10 s episodes. PPO collects 24 steps per environment and performs five epochs with four minibatches. Both curricula use clipping 0.2, discount 0.99, and GAE parameter 0.95; ACC/WFC use learning rates $2\times10^{-4}/1.5\times10^{-4}$ and entropy coefficients 0.003/0.004. Motion and dynamics variation follow separate curricula. All amplitude–frequency, recognized-input, dynamic phase-ablation, and capability-loss evaluations freeze the WFC policy at iteration 39,999.

### *B. Randomization and Deployment Interface*

Per-environment randomization covers static/dynamic friction in [0.60,1.40]/[0.50,1.20], restitution in [0,0.10], torso center-of-mass offsets, motor zero offsets, external pushes, IMU/encoder bias and noise, and 0–2-step sensing delay. Each foot is represented by six independent contact pads to reduce simulator dependence of large-area contact.

## VIII. Results and Evaluation Protocol

### *A. Human Gait Kinematics Reconstruction*

Five synchronized walking recordings contain 21,772 raw samples at approximately 19.75 Hz. Two IMUs provide 18 input channels, paired with hip, knee, and ankle angle targets.

Across all 4,267 test windows, the estimator reaches 3.242° MAE, 5.607° RMSE, and $R^2 = 0.819$. The results are shown in Table I. Versus 9.754°, 13.280°, and −0.016 for a training-mean baseline: reductions of 66.8% in MAE and 57.8% in RMSE. Hip, knee, and ankle correlations are 0.924, 0.916, and 0.708. The results are shown in Figure 5. and TABLE I.

### *B. Factorized Multi Trajectory Tracking*

The 17 trajectories are amplitude variants, each is evaluated at seven commanded cadence scales (shown in TABLE II. ), producing 119 conditions spanning 0.484–1.548 cycles/s. Each condition starts at frame offsets 0, 20, and 41 within the 62-frame cycle: three starting phases, not additional frequencies. The resulting 357 trials run for up to 10 s; tracking statistics exclude the first second. Sensor, dynamics, and reset randomization are disabled to isolate reference variation.

ACC varies both amplitude and cadence, whereas WFC trains only nominal-amplitude variant 0 over the wider cadence interval. Thus, non-nominal rows in Fig. 6 test amplitude transfer of the final WFC policy, not amplitude conditions sampled during WFC training. This distinction prevents confusing evaluation coverage with training exposure. The frozen WFC policy completes 356/357 trials (99.72%), including all 21 amplitude trials and all 84 trials for ACC-held-out IDs 13–16. Completed-trial bilateral hip RMSE averages 3.66° with correlation 0.953. Condition-wise means span 2.59–5.68° and 0.917–0.973. One policy therefore tracks the dominant hip waveform across the tested group, without retraining per condition. For variant 0, increasing cadence from 0.50× through 1.00× to 1.60× changes bilateral hip RMSE from 3.23° through 3.34° to 4.63°; the realized/reference amplitude ratio drops from 0.84 to 0.71. TABLE III. aggregates all 17 variants at each frequency. Together with Figure 7, these results characterize cadence-dependent tracking and realized excursion.

Completion counts all 357 trials; errors average the 356 completed trials. The sole termination is at 1.12 s for variant 9 (hip/knee/ankle scales 1.121/1.166/0.917), 1.60× cadence, and starting phase zero; the other two starts complete. TABLE III. lists commanded frequencies; this grid uses WFC with the original ±10% RRC. The grid characterizes tracking across amplitudes and frequencies under nominal dynamics. Hip, knee, and ankle RMSE are 3.66°, 6.33°, and 5.93°, respectively; knee correlation is 0.272.

Figure 8. presents time-matched snapshots of the kinematically replayed human reference and the policy-controlled exoskeleton. Their alternating swing and stance poses make the intended gait coordination visually interpretable. The human mannequin follows prescribed kinematics, whereas the exoskeleton moves under simulated dynamics; quantitative phase and joint-angle agreement are assessed separately using the time-series results.

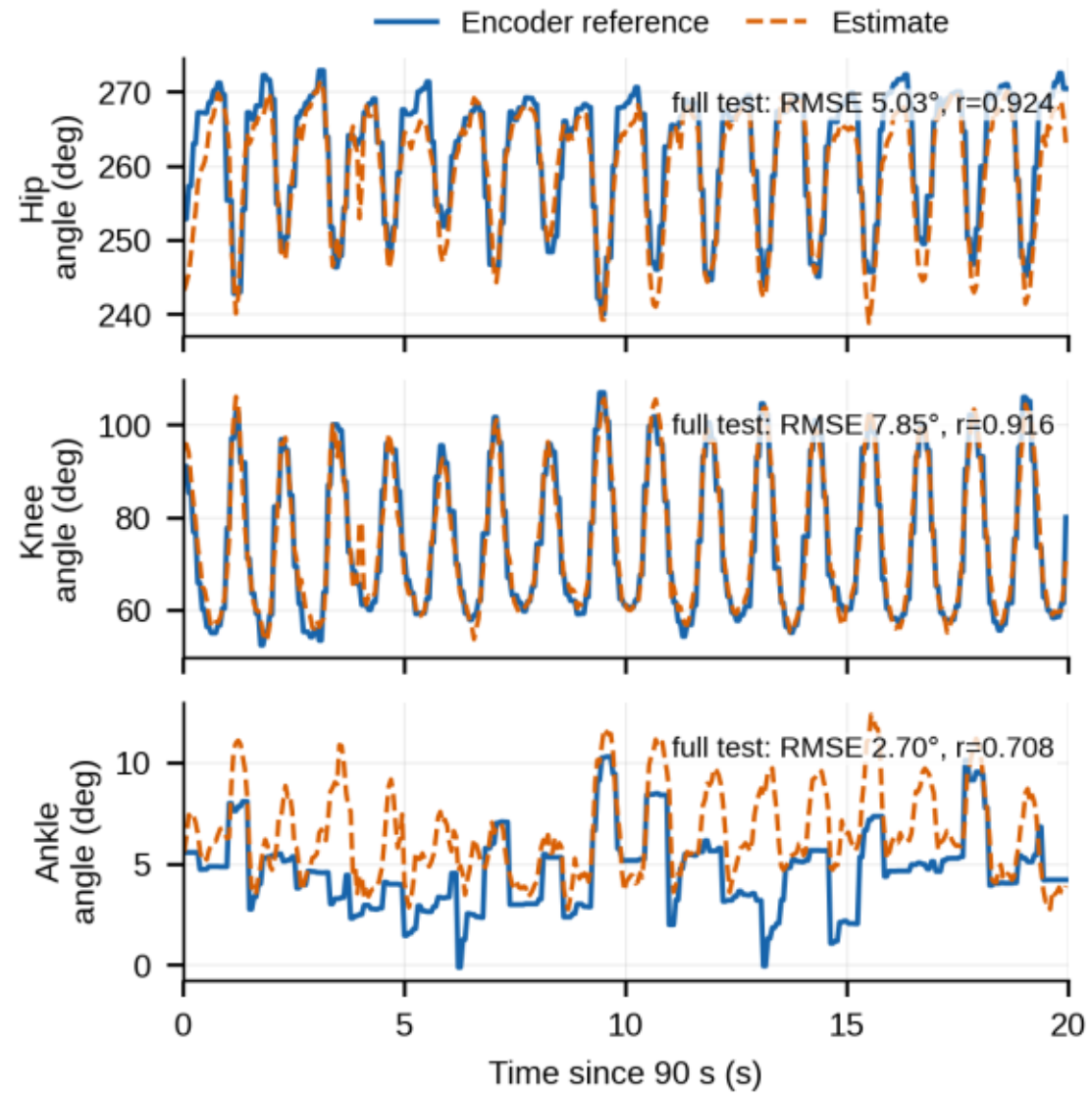

Figure 5. Results of Human gait reconstruction.

TABLE I. Results of Human Joint Angle Reconstruction

| Joint | MAE° | RMSE° | $R^2$ | Corr. | Baseline RMSE° |
|---|---|---|---|---|---|
| Hip | 3.76 | 5.03 | 0.808 | 0.924 | 11.66 |
| Knee | 4.00 | 7.85 | 0.836 | 0.916 | 19.41 |
| Ankle | 1.97 | 2.70 | 0.437 | 0.708 | 4.07 |
| Overall | 3.24 | 5.61 | 0.819 | — | 13.28 |

TABLE II. Amplitude and Cadence Factors

| Factor | ACC train | WFC train | Evaluation |
|---|---|---|---|
| Hip scale | 0.85–1.15 | 1.00 | 0.85–1.15 |
| Knee scale | 0.80–1.20 | 1.00 | 0.80–1.20 |
| Ankle scale | 0.90–1.10 | 1.00 | 0.90–1.10 |
| Variant IDs | 0–12 | 0 | 0–16 (17) |
| Cadence × | 0.75–1.25 | 0.50–1.60 | 7 levels* |

* Cadence levels: 0.50, 0.65, 0.80, 1.00, 1.20, 1.40, 1.60. Every amplitude variant is crossed with all seven levels.

### *C. PhaseSync Exo Method*

We evaluate HCA through two complementary dynamic tests: maintaining consistency with the human clock, and acquiring human phase from an initially misaligned reference. Both freeze the same WFC policy, use 2 s common warmup followed by 20 s evaluation, and retain the original physics and termination criteria. Robot phase is estimated by an independent whole-cycle bilateral-hip observer and compared with the human clock; the adapter's own cursor is not used as the robot-phase measurement. Noise/randomization is disabled and adapter gains remain unchanged. Table IV summarizes the comparisons; Fig. 9 shows their temporal behavior.

Human-clock consistency. FC, robot-relative correction (RRC), and HCA are compared over ten starts, two common source/reference phase shifts (±0.15 cycle), and two cadence steps (0.9×/1.1×): $N = 10 \times 2 \times 2 = 40$ conditions per rule. Here the human and delivered clocks coincide at onset. HCA reduces RRC's human-clock phase MAE by 88.5% (95.55° to 10.99°), while delivered-reference hip RMSE stays close (3.36°/3.42°). Mean final reference lag falls from 0.905 to 0.018 cycle. The improvement therefore comes from suppressing reference drift, rather than changing the learned tracking policy. FC gives 9.74° MAE in this already aligned condition, providing the nominal clock-following baseline.

Phase reacquisition. The ongoing reference instead starts ±0.15 or ±0.25 cycle away from an independent human source, at 0.9×/1.1× cadence: $N = 10 \times 4 \times 2 = 80$ matched conditions per rule, plus 20 zero-offset shams each. FC continuation keeps the correct cadence without changing its initial phase. HCA reduces source-phase MAE by 80.7% (72.02° to 13.90°), and steady MAE by 84.8% (72.02° to 10.98°). Human-trajectory hip RMSE decreases from 10.29° to 4.00°; all 80 paired conditions improve in both phase MAE and human-trajectory hip RMSE. Thus matching cadence alone does not correct initial timing disagreement, whereas human anchoring reacquires the source Continuity during acquisition. A stronger FC-reset control assigns the human phase immediately, achieving 9.70° phase MAE but producing reference jumps in all 80 trials, including 40 backward assignments. HCA avoids these jumps and reduces 0–2 s delivered-reference hip RMSE by 22.7% (4.23° to 3.27°; 69/80 pairs improve). All 420 rollouts across the two studies complete without falls or tracking termination. Together, the results demonstrate HCA's two benefits: maintaining human-clock consistency during adaptation and recovering phase smoothly when a fixed-rate reference starts misaligned.

The supplement retains the kinematic signal test, recognized-input joint trajectories, and actuator-derating results. Derating confirms continued walking but not a uniform accuracy or recovery gain; the advantage established here concerns reference-clock consistency and continuous phase acquisition.

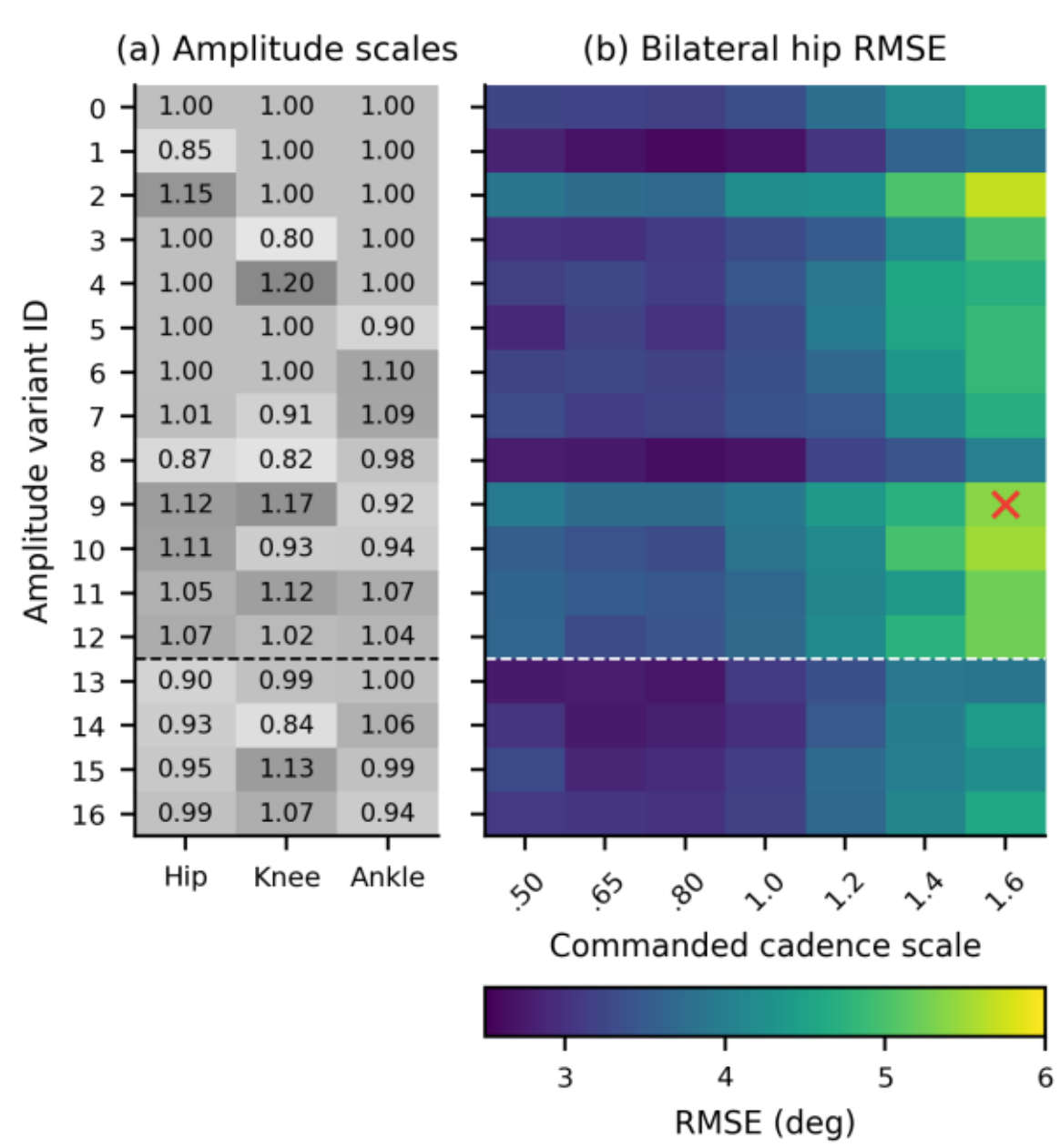

Figure 6. Amplitude–frequency grid. (a) Hip/knee/ankle scales define each row. (b) Columns vary cadence; color shows bilateral hip angle RMSE, averaged over completed trials at three starting phases—not phase error.

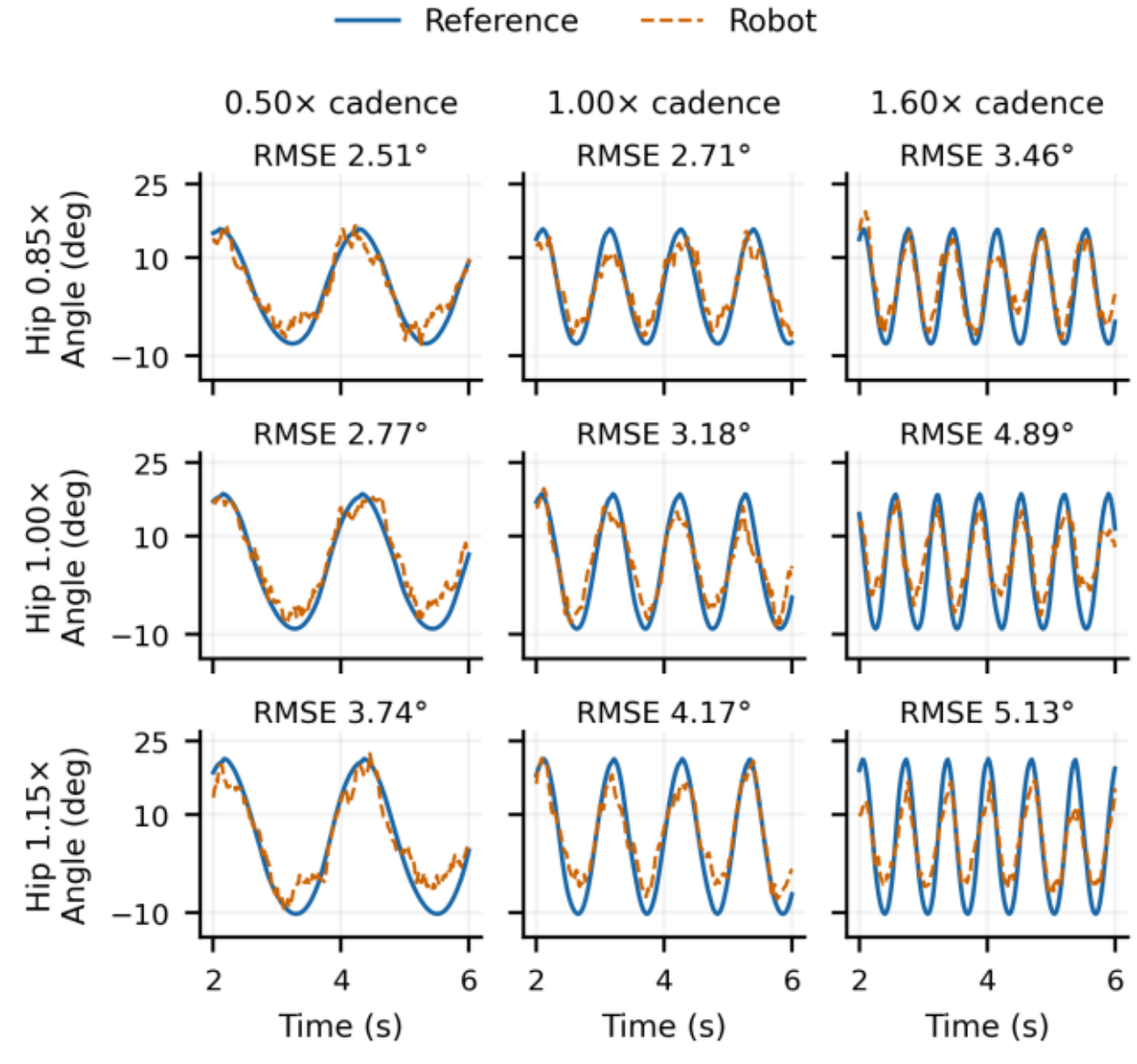

Figure 7. Responses at low, nominal, and high amplitude–cadence settings.

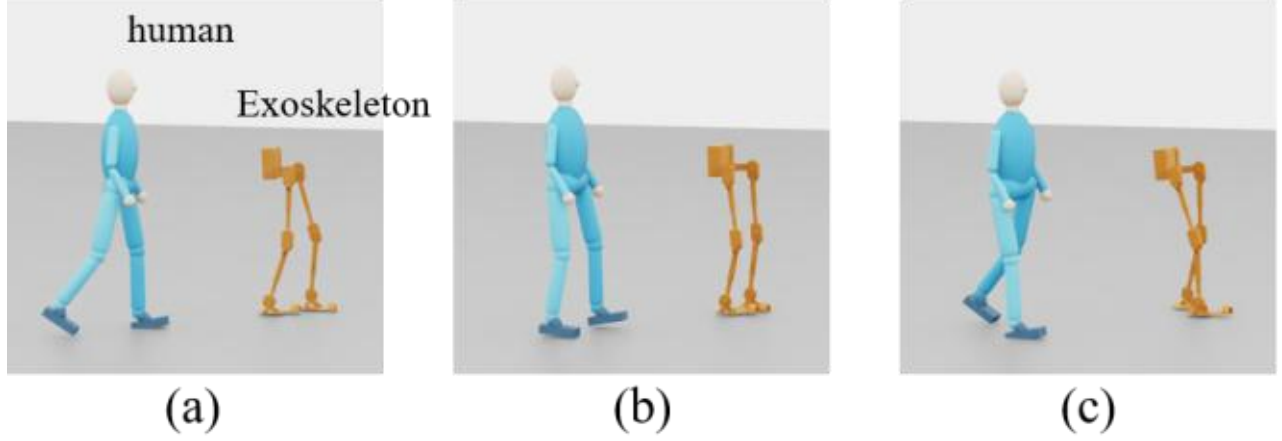

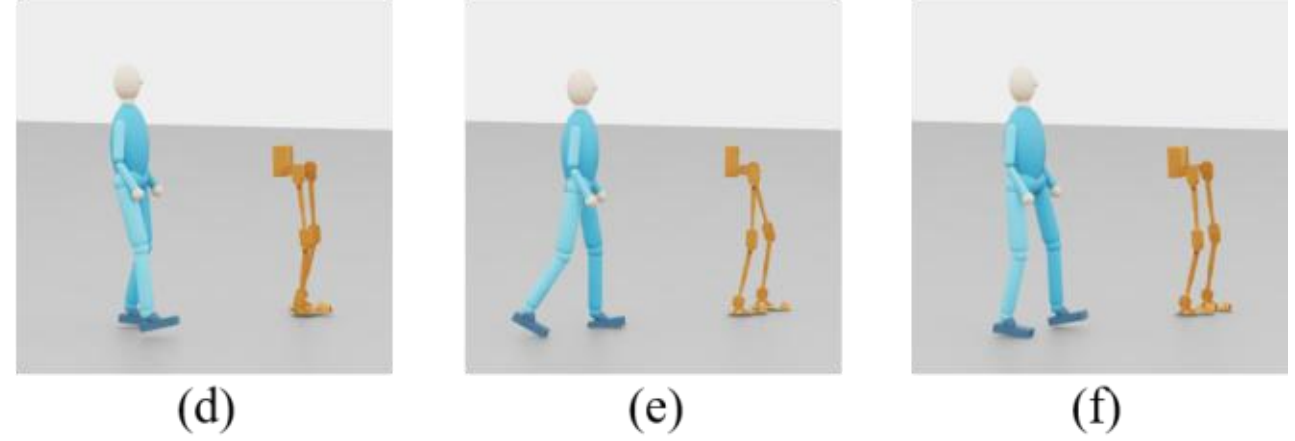

(a) (b) (c) (d) (e) (f)

Figure 8. Kinematic human reference and policy-controlled exoskeleton at matched times; the mannequin does not undergo dynamics simulation.

TABLE III. TRACKING ACROSS THE COMMANDED FREQUENCY GRID

| Scale | Cycle Hz | Done | Hip RMSE ° | Knee RMSE ° | Ankle RMSE ° |
|---|---|---|---|---|---|
| 0.50 | 0.484 | 51/51 | 3.25 | 6.10 | 6.14 |
| 0.65 | 0.629 | 51/51 | 3.15 | 5.97 | 6.11 |
| 0.80 | 0.774 | 51/51 | 3.12 | 6.06 | 6.02 |
| 1.00 | 0.968 | 51/51 | 3.36 | 6.38 | 5.99 |
| 1.20 | 1.161 | 51/51 | 3.75 | 6.55 | 5.87 |
| 1.40 | 1.355 | 51/51 | 4.29 | 6.61 | 5.72 |
| 1.60 | 1.548 | 50/51 | 4.73 | 6.68 | 5.67 |
| All | — | 356/357 | 3.66 | ` | 5.93 |

TABLE IV. HUMAN CLOCK CONSISTENCY AND PHASE REACQUISITION IN INITIALLY COINCIDENT CLOCKS

| Metric | FC | RRC | HCA |
|---|---|---|---|
| Human-clock phase MAE (°) | 9.74 | 95.55 | 10.99 |
| Hip RMSE to human (°) | 3.37 | 13.00 | 3.44 |
| Hip RMSE to delivered ref. (°) | 3.37 | 3.36 | 3.42 |
| Mean final reference lag (cycle) | 0.000 | 0.905 | 0.018 |
| Completed / falls | 40 / 0 | 40 / 0 | 40 / 0 |

TABLE V. HUMAN CLOCK CONSISTENCY AND PHASE REACQUISITION IN INITIALLY MISALIGNED CLOCKS

| Metric | FC cont. | FC reset | HCA |
|---|---|---|---|
| Human-clock phase MAE (°) | 72.02 | 9.70 | 13.90 |
| Steady human phase MAE (°) | 72.02 | 9.35 | 10.98 |
| Hip RMSE to human (°) | 10.29 | 3.39 | 4.00 |
| 0–2 s hip RMSE to ref. (°) | 3.29 | 4.23 | 3.27 |
| Onset jumps / backward jumps | 0 / 0 | 80 / 40 | 0 / 0 |
| Completed / falls | 80 / 0 | 80 / 0 | 80 / 0 |

## IX. CONCLUSION

PhaseSync-Exo connects wearable gait reconstruction and multi-trajectory dynamics learning through a human-clock-anchored reference interface. Reconstruction achieves 3.24° MAE on a held-out recording, and the frozen WFC policy completes 356/357 amplitude–frequency trials. The two dynamic timing studies establish complementary benefits: HCA suppresses the drift of RRC and recovers phase offsets that FC continuation retains, reducing human-clock phase MAE by 88.5% and 80.7%, respectively. It also lowers 0–2 s delivered-reference hip RMSE by 22.7% relative to immediate reset, without reference jumps. All 420 timing rollouts complete without falls. Together, these results support human-clock consistency and continuous phase acquisition without retraining the dynamics policy, providing a simulation basis for subsequent physical human–exoskeleton validation.

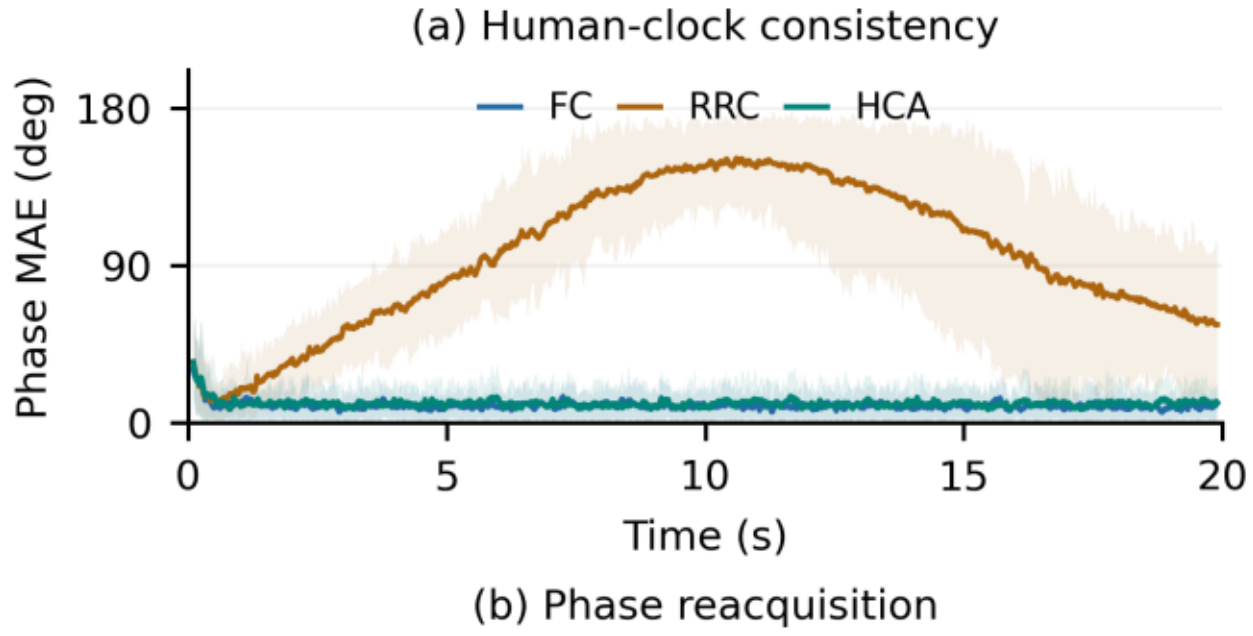

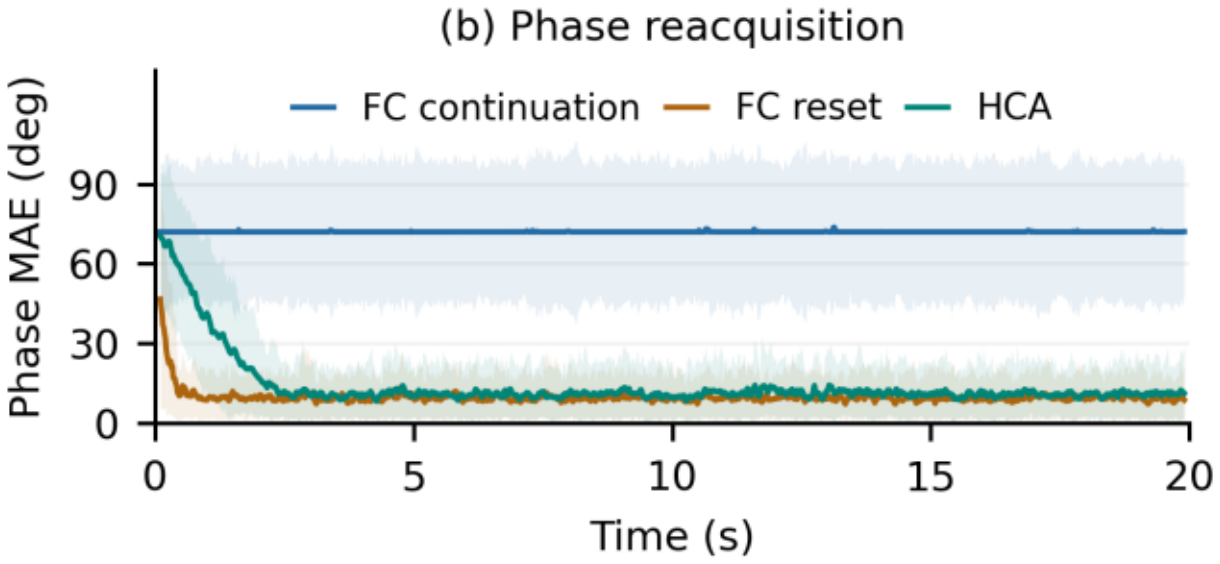

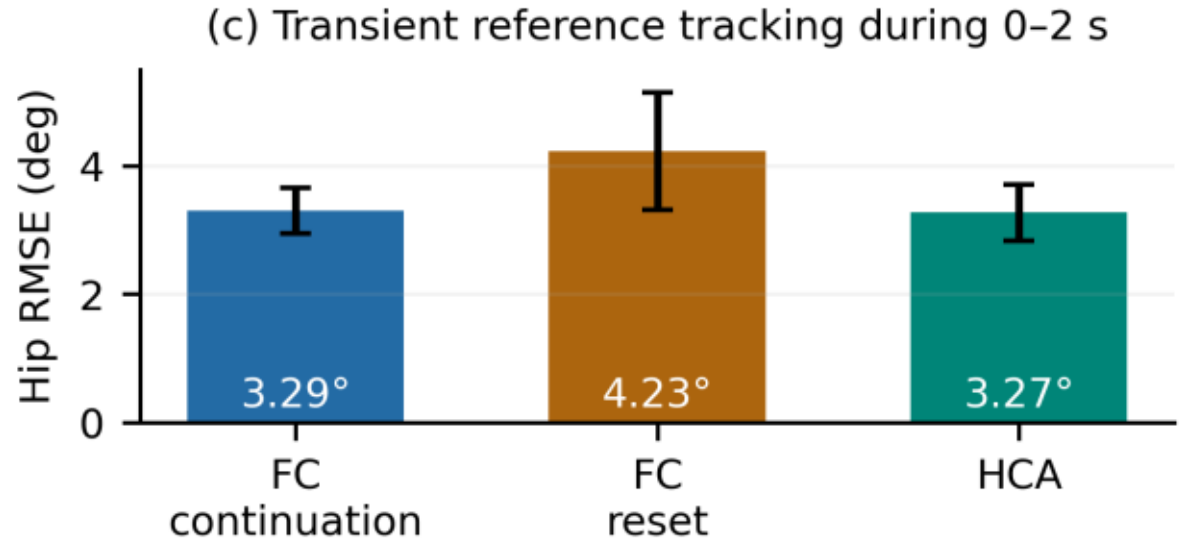

Figure 9. Two frozen-policy dynamic comparisons. (a) Human-clock consistency, 40 trials per rule. (b) Phase reacquisition, 80 nonzero-offset trials per rule. Curves show mean absolute circular phase error; shading is the 10th–90th trial percentile. (c) Acquisition-window delivered-reference hip RMSE, mean ± SD. FC reset is an instantaneous phase assignment, distinct from fixed-rate continuation.

## REFERENCES

[1] X. B. Peng, P. Abbeel, S. Levine, and M. van de Panne, "DeepMimic: Example-guided deep reinforcement learning of physics-based character skills," *ACM Trans. Graph.*, vol. 37, no. 4, 2018, doi: 10.1145/3197517.3201311.

[2] X. B. Peng, Z. Ma, P. Abbeel, S. Levine, and A. Kanazawa, "AMP: Adversarial motion priors for stylized physics-based character control," *ACM Trans. Graph.*, vol. 40, no. 4, 2021, doi: 10.1145/3450626.3459670.

[3] Q. Liao, T. E. Truong, X. Huang, Y. Gao, G. Tevet, K. Sreenath, and C. K. Liu, "BeyondMimic: From motion tracking to versatile humanoid control via guided diffusion," *arXiv*:2508.08241, 2025, doi: 10.48550/arXiv.2508.08241.

[4] J. P. Araujo, Y. Ze, P. Xu, J. Wu, and C. K. Liu, "Retargeting matters: General motion retargeting for humanoid motion tracking," *arXiv*:2510.02252, 2025.

[5] J. Schulman, F. Wolski, P. Dhariwal, A. Radford, and O. Klimov, "Proximal policy optimization algorithms," *arXiv*:1707.06347, 2017.

[6] J. Schulman, P. Moritz, S. Levine, M. Jordan, and P. Abbeel, "High-dimensional continuous control using generalized advantage estimation," *arXiv*:1506.02438, 2015.

[7] N. Rudin, D. Hoeller, P. Reist, and M. Hutter, "Learning to walk in minutes using massively parallel deep reinforcement learning," in *Proc. Conf. Robot Learning*, 2022, pp. 91–100.

[8] X. B. Peng, M. Andrychowicz, W. Zaremba, and P. Abbeel, “Sim-to-real transfer of robotic control with dynamics randomization,” in *Proc. IEEE Int. Conf. Robot. Autom.*, 2018, pp. 3803–3810.

[9] S. Luo, G. Androwis, S. Adamovich, H. Su, and X. Zhou, “Reinforcement learning and control of a lower extremity exoskeleton for squat assistance,” *arXiv*:2105.03489, 2021.

[10] Y. Shen, Y. Wang, Z. Zhao, C. Li, and M. Q.-H. Meng, “A modular lower limb exoskeleton system with RL based walking assistance control,” in *Proc. IEEE Int. Conf. Robot. Biomimetics*, 2021, pp. 1258–1263.

[11] X. Wang, Y. Ma, C. Chen, Z. Wang, S. Guo, K. H. Low, and X. Wu, “Effective prediction of gait phase for assisted walking by means of gait-based adaptive oscillators,” *IEEE Trans. Autom. Sci. Eng.*, vol. 22, pp. 10205–10215, 2025, doi: 10.1109/TASE.2024.3520148.

[12] M. Shushtari, H. Dinovitzer, J. Weng, and A. Arami, “Ultra-robust real-time estimation of gait phase,” *IEEE Trans. Neural Syst. Rehabil. Eng.*, vol. 30, pp. 2793–2801, 2022, doi: 10.1109/TNSRE.2022.3207919.

[13] M. Shushtari and A. Arami, “Human–exoskeleton disagreement resolution through interaction torque minimization: Experimental results,” in Proc. *IEEE RAS/EMBS Int. Conf. Biomed. Robot*. Biomechatronics, 2024, pp. 9–14, doi: 10.1109/BioRob60516.2024.10719865.